\documentclass[11pt,a4paper]{article}
\usepackage[hyperref]{naaclhlt2018}
\usepackage{times}
\usepackage{latexsym}
\usepackage{amsmath}
\usepackage{url}
\usepackage{verbatim}
\usepackage{wrapfig}
\usepackage{graphicx}
\usepackage{amsfonts}
\usepackage{cleveref}
\usepackage{subcaption}
\usepackage{url}
\aclfinalcopy 

\title{SHAPED: Shared-Private Encoder-Decoder for Text Style Adaptation}

\author{Ye Zhang\thanks{\;\;Work done as an intern at Google AI.} \\
  UT Austin \\
  {\tt yezhang@utexas.edu} \\\And
  Nan Ding \;\;\;\;\;\;\;\;\;\; Radu Soricut\\
  Google AI \\
  {\tt \{dingnan,rsoricut\}@google.com} }
\begin{document}

\maketitle
\begin{abstract}
Supervised training of abstractive language generation models results in learning conditional probabilities over language sequences based on the supervised training signal. When the training signal contains a variety of writing styles, such models may end up learning an 'average' style that is directly influenced by the training data make-up and cannot be controlled by the needs of an application. We describe a family of model architectures capable of capturing both generic language characteristics via shared model parameters, as well as particular style characteristics via private model parameters. Such models are able to generate language according to a specific learned style, while still taking advantage of their power to model generic language phenomena. Furthermore, we describe an extension that uses a mixture of output distributions from all learned styles to perform on-the-fly style adaptation based on the textual input alone. Experimentally, we find that the proposed models consistently outperform models that encapsulate single-style or average-style language generation capabilities.
\end{abstract}

\section{Introduction}
Encoder-decoder models have recently pushed forward the state-of-the-art performance on a variety of language generation tasks, including machine translation~\citep{bahdanau2015neural,wu2016gnmt,vaswani2017attention}, text summarization~\citep{rush2015neural,nallapati2016abstractive, see2017get}, dialog systems~\citep{li2016deep, asghar2017deep}, and image captioning~\citep{xu2015show,mixer15,liu2017optimization}.
This framework consists of an encoder that reads the input data and encodes it as a sequence of vectors, which is in turn used by a decoder to generate another sequence of vectors used to produce output symbols step by step.

The prevalent approach to training such a model is to update all the model parameters using all the examples in the training data (over multiple epochs).
This is a reasonable approach, under the assumption that we are modeling a single underlying distribution in the data.
However, in many applications and for many natural language datasets, there exist multiple underlying distributions, characterizing a variety of language styles.
For instance, the widely-used Gigaword dataset~\citep{graff2003english} consists of a collection of articles written by various publishers (The New York Times, Agence France Presse, Xinhua News, etc.), each with its own style characteristics.
Training a model's parameters on all the training examples results in an averaging effect across style characteristics, which may lower the quality of the outputs; additionally, this averaging effect may be completely undesirable for applications that require a level of control over the output style.
At the opposite end of the spectrum, one can choose to train one independent model per each underlying distribution (assuming we have the appropriate signals for identifying them at training time).
This approach misses the opportunity to exploit common properties shared by these distributions (e.g., generic characteristics of a language, such as noun-adjective position), and leads to models that are under-trained due to limited data availability per distribution.

In order to address these issues, we propose a novel neural architecture called
SHAPED (\underline{sha}red-\underline{p}rivate \underline{e}ncoder-\underline{d}ecoder).
This architecture has both shared encoder/decoder parameters that are updated based on all the training examples, as well as private encoder/decoder parameters that are updated using only examples from their corresponding underlying training distributions.
In addition to learning different parametrization between the shared model and the private models, we jointly learn a classifier to estimate the probability of each example belonging to each of the underlying training distributions.
In such a setting, the shared parameters ('shared model') are expected to learn characteristics shared by the entire set of training examples (i.e., language generic), whereas each private parameter set ('private model') learns particular characteristics (i.e., style specific) of their corresponding training distribution.
At the same time, the classifier is expected to learn a probability distribution over the labels used to identify the underlying distributions present in the input data.
At test time, there are two possible scenarios.
In the first one, the input signal explicitly contains information about the underlying distribution (e.g., the publisher's identity).
In this case, we feed the data into the shared model and also the corresponding private model, and perform sequence generation based on a concatenation of their vector outputs; we refer to this model as the SHAPED model.
In a second scenario, the information about the underlying distribution is either not available, or it refers to a distribution that was not seen during training.
In this case, we feed the data into the shared model and all the private models; the output distribution of the symbols of the decoding sequence is estimated using a mixture of distributions from all the decoders, weighted according to the classifier's estimates for that particular example; we refer to this model as the Mix-SHAPED model.

We test our models on the headline-generation task based on the aforementioned Gigaword dataset.
When the publisher's identity is presented as part of the input, we show that the SHAPED model significantly surpasses the performance of the shared encoder-decoder baseline, as well as the performance of private models (where one individual, per-publisher model is trained for each in-domain style).
When the publisher's identity is not presented as part of the input (i.e., not presented at run-time but revealed at evaluation-time for measurement purposes), we show that the Mix-SHAPED model exhibits a high level of classification accuracy based on textual inputs alone (accuracy percentage in the 80s overall, varying by individual publisher), while its generation accuracy still surpasses the performance of the baseline models.
Finally, when the publisher's identity is unknown to the model (i.e., a publisher that was not part of the training dataset), we show that the Mix-SHAPED model performance far surpasses the shared model performance, due to the ability of the Mix-SHAPED model to perform on-the-fly adaptation of output style.
This feat comes from our model's ability to perform two distinct tasks: match the incoming, previously-unseen input style to existing styles learned at training time, and use the correlations learned at training time between input and output style characteristics to generate style-appropriate token sequences.

\section{Related Work}
\label{sec:relate}
\subsection*{Encoder-Decoder Models for Structured Output Prediction}
Encoder-decoder architectures have been successfully applied to a variety of structure prediction tasks recently.
Tasks for which such architectures have achieved state-of-the-art results include machine translation~\citep{bahdanau2015neural,wu2016gnmt, vaswani2017attention}, automatic text summarization~\citep{rush2015neural, chopra2016abstractive, nallapati2016abstractive, paulus2017deep, nema2017diversity},
sentence simplification~\citep{filippova2015sentence, zhang2017sentence},
dialog systems~\citep{li2016deep, li2017adversarial, asghar2017deep},
image captioning~\citep{vinyals2015show,xu2015show,mixer15,liu2017optimization}, etc.
By far the most used implementation of such architectures is based on the original sequence-to-sequence model~\citep{sutskever2014sequence}, augmented with its attention-based extension~\cite{bahdanau2015neural}.
Although our SHAPED and Mix-SHAPED model formulations do not depend on a particular architecture implementation, we do make use of the~\cite{bahdanau2015neural} model to instantiate our models.

\subsection*{Domain Adaptation for Neural Network Models}
One general approach to domain adaptation for natural language tasks is to perform data/feature augmentation that represents inputs as both general and domain-dependent data, as originally proposed in~\citep{daume2009frustratingly}, and ported to neural models in~\citep{kim2016frustratingly}.
For computer vision tasks, a line of work related to our approach has been proposed by~\citet{bousmalis2016domain} using what they call domain separation networks.
As a tool for studying unsupervised domain adaptation for image recognition tasks, their proposal uses CNNs for encoding an image into a feature representation, and also for reconstructing the input sample.
It also makes use of a private encoder for each domain, and a shared encoder for both the source and the target domain.
The approach we take in this paper shares this idea of model parametrization according to the domain/style,
but goes further with the Mix-SHAPED model, performing on-the-fly adaptation of the model outputs.
Other CNN-based domain adaptation methods for object recognition tasks are presented in~\cite{long2016unsupervised, chopra2013dlid, tzeng2015simultaneous,sener2016learning}.

For NLP tasks, \citet{peng2017multi} take a multi-task approach to domain adaptation and sequence tagging.
They use a shared encoder to represent instances from all of the domains, and use a domain projection layer to project the shared layer into a domain-specific space.
They only consider the supervised domain-adaptation case, in which labeled training data exists for the target domain.
\citet{glorot2011domain} use auto-encoders for learning a high-level feature extraction across domains for sentiment analysis, while~\citet{zhou2016bi} employ auto-encoders to directly transfer the examples across different domains also for the same sentiment analysis task.
\citet{hua2017pilot} perform an experimental analysis on domain adaptation for neural abstractive summarization.

An important requirement of all the methods in the related work described above is that they require access to the (unlabeled) target domain data, in order to learn a domain-invariant representation across source and target domains.
In contrast, our Mix-SHAPED model does not need access to a target domain or style at training time, and instead performs the adaptation on-the-fly, according to the specifics of the input data and the correlations learned at training time between available input and output style characteristics.
As such, it is a more general approach, which allows adaptation for a much larger set of target styles, under the weaker assumption that there exists one or more styles present in the training data that can act as representative underlying distributions.

\section{Model Architecture}
\label{sec:red}
Generally speaking, a standard encoder-decoder model has two components: an encoder that takes as input a sequence of symbols
$\mathbf{x}=(x_1,x_2,...,x_{T_x})$ and encodes them into a set of vectors $\mathbf{H}=(h_1,h_2,...,h_{T_x})$,
\begin{equation}
\mathbf{H} = f_{\text{enc}}(\mathbf{x}),
\end{equation}
\noindent
where $f_{\text{enc}}$ is the computation unit in the encoder;
and, a decoder that generates output symbols at each time stamp $t$, conditioned on $\mathbf{H}$ as well as the decoder inputs $\mathbf{y}_{1:t-1}$,
\begin{equation}
s_t = f_{\text{dec}}(\mathbf{y}_{1:t-1},\mathbf{H}),
\end{equation}
\noindent
where $f_{\text{dec}}$ is the computation unit in the decoder.
Instantiations of this framework include the widely-used attention-based sequence-to-sequence model~\citep{bahdanau2015neural}, in which $f_{\text{enc}}$ and $f_{\text{dec}}$ are implemented by an RNN architecture using LSTM~\citep{hochreiter1997long} or GRU~\citep{chung2014empirical} units.
A more recent instantiation of this architecture is the Transformer model~\citep{vaswani2017attention}, built using self-attention layers.

\subsection{SHAPED: Shared-private encoder-decoder}
The abstract encoder-decoder model described above is usually trained over all examples in the training data.
We call such a model a \emph{shared} encoder-decoder model, because the model parameters are shared across all training and test instances.
Formally, the shared encoder-decoder consists of the computation units $f^s_\text{enc}$ and $f^s_\text{dec}$.
Given an instance $\mathbf{x}$, it generates a sequence of vectors $\mathbf{S}^s = (s_1^s,...s_T^s)$ by:
\begin{equation}
\label{Eq:shared_network}
\mathbf{H}^s = f_{\text{enc}}^s(\mathbf{x}), s_t^s = f_{\text{dec}}^s(\mathbf{y}_{1:t-1},\mathbf{H}^s).
\end{equation}

\begin{figure}[h]
      \includegraphics[width=0.45\textwidth]{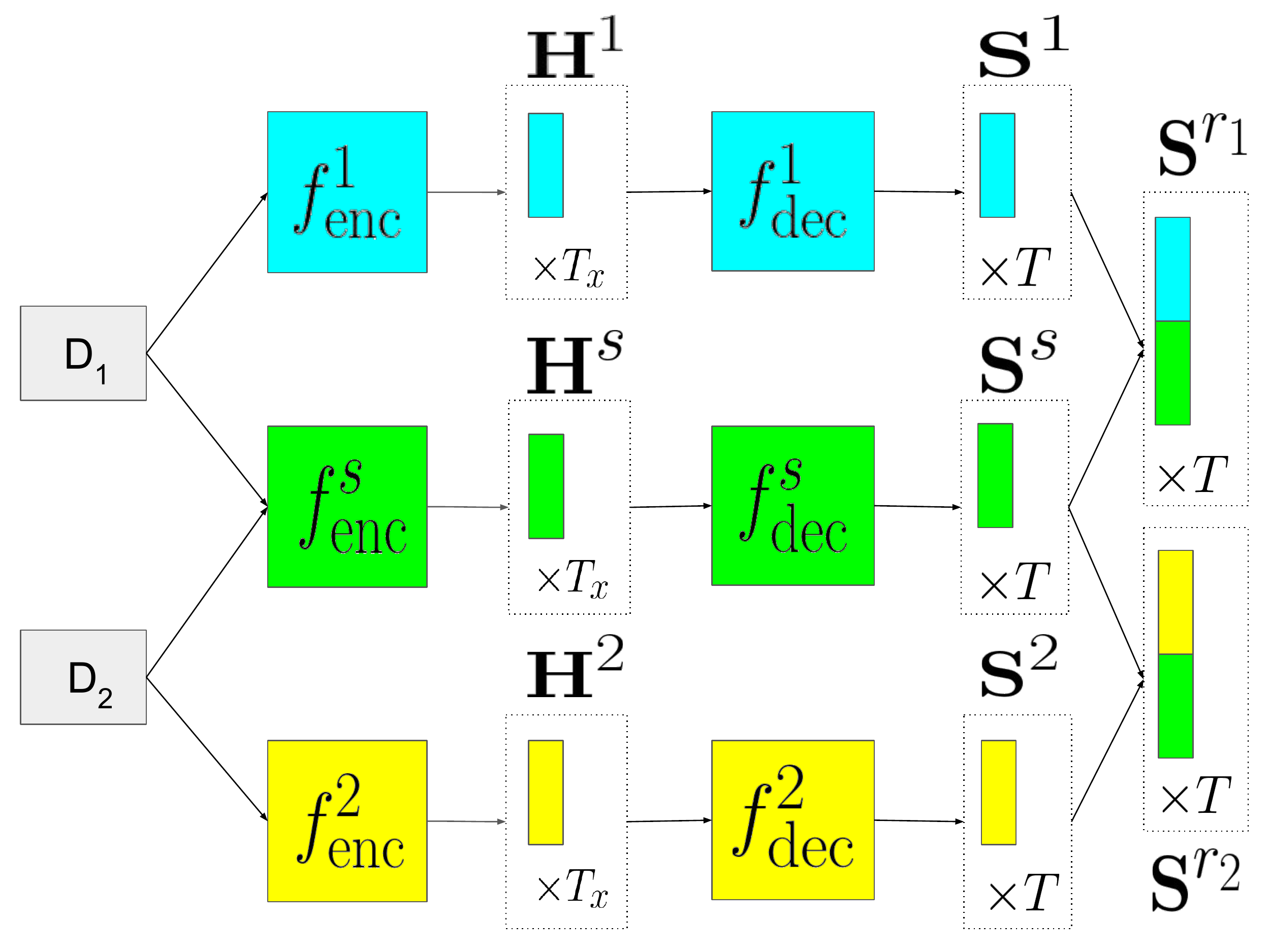}
 \caption{Illustration of the SHAPED model using two styles $D_1$ and $D_2$. $D_1$ articles pass through the private encoder $f_{\text{enc}}^1$ and decoder $f_{\text{dec}}^1$. $D_2$ articles pass through the private encoder $f_{\text{enc}}^2$ and decoder $f_{\text{dec}}^2$. Both of them also go through the shared encoder $f_{\text{dec}}^s$ and decoder $f_{\text{dec}}^s$.}
  \label{Res-RNN}
  \end{figure}

The drawback of the shared encoder-decoder is that it fails to account for particular properties of each style that may be present in the data.
In order to capture such particular style characteristics, a straightforward solution is to train a \emph{private} model for each style.
Assuming a style set $\mathbb{D}=\{D_1, D_2...,D_{|\mathbb{D}|}\}$, such a solution implies that each style has its own private encoder computation unit and decoder computation unit.
At both training and testing time, each private encoder and decoder only process instances that belong to their own style.
Given an instance along with its style ($\mathbf{x}$, $z$) where $z \in \{1, \ldots, |\mathbb{D}|\}$, the private encoder-decoder generates a sequence of vectors $\mathbf{S}^z = (s_1^z,...s_T^z)$ by:
\begin{equation}
\label{Eq:private_encoder}
\mathbf{H}^z = f_{\text{enc}}^z(\mathbf{x}), s_t^z = f_{\text{dec}}^z(\mathbf{y}_{1:t-1},\mathbf{H}^z).
\end{equation}
Although the private encoder/decoder models do preserve style characteristics, they fail to take into account the common language features shared across styles.
Furthermore, since each style is represented by a subset of the entire training set, such private models may end up as under-trained, due to limited number of available data examples.

In order to efficiently capture both common and unique features of data with different styles, we propose the SHAPED model.
In the SHAPED model, each data-point goes through both the shared encoder-decoder and its corresponding private encoder-decoder.
At each step of the decoder, the output from private and shared ones are concatenated to form a new vector:
\begin{equation}
\label{Eq:concat}
s_t^{r_z}=[s_t^z, s_t^s],
\end{equation}
that contains both private features for style $z$ and shared features induced from other styles, as illustrated in Fig \ref{Res-RNN}.
The output symbol distribution over tokens $o_t\in V$ (where $V$ is the output vocabulary) at step $t$ is given by:
 \begin{equation}
 \label{Eq:ProjSoftmax}
 p(o_t|\mathbf{x},y_{1:t-1},z) = \text{Softmax}(g(s_t^{r_z})),
 \end{equation}
where $g$ is a multi-layer feed-forward network that maps $s_t^{r_z}$ to a vector of size $|V|$.
Given $N$ training examples $(\mathbf{x}^{(1)},\mathbf{y}^{(1)}, z^{(1)}),\ldots,(\mathbf{x}^{(N)},\mathbf{y}^{(N)}, z^{(N)})$, the conditional probability of the output $\mathbf{y}^{(i)}$ given article $\mathbf{x}^{(i)}$ and its style $z^{(i)} \in \{1, \ldots, |\mathbb{D}|\}$ is:
 \begin{equation}
 p(\mathbf{y}^{(i)}|\mathbf{x}^{(i)},z^{(i)})=\prod_t p(o_t=y_{t}^{(i)}|\mathbf{x}^{(i)},\mathbf{y}_{1:t-1}^{(i)},z^{(i)}).
 \end{equation}
At inference time, given an article $\mathbf{x}$ with style $z$, we feed $\mathbf{x}$ into $f^s_\text{enc}, f^s_\text{dec}, f^z_\text{enc}, f^z_\text{dec}$ (Eq.~\ref{Eq:shared_network}-\ref{Eq:private_encoder}) and obtain symbol distributions at each step $t$ using Eq.~\ref{Eq:ProjSoftmax}.
We sample from the distribution and obtain a symbol $o_t$ which will be used as the estimated $y_{t}$ and fed to the next steps.

\subsection{The Mix-SHAPED Model}
One limitation of the above model is that it can only handle test data containing an explicit style label from $\mathbb{D} = \{D_1, D_2...,D_{|\mathbb{D}|}\}$.
However, there is frequently the case that, at test time, the style label is not present as part of the input, or that the input style is not part of the modeled set $\mathbb{D}$.

\begin{figure}[!h]
 \begin{center}
      \includegraphics[width=0.45\textwidth]{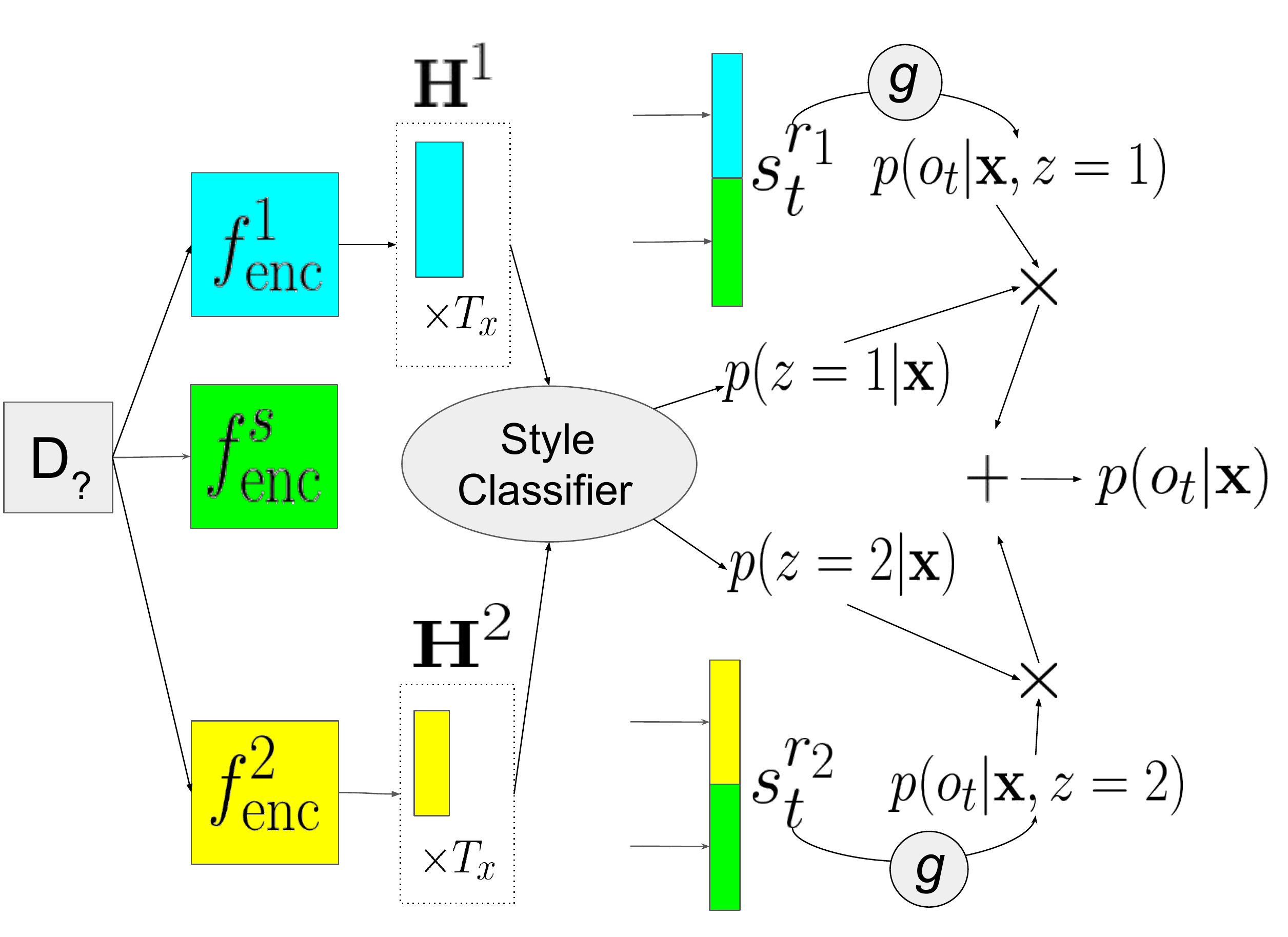}
  \end{center}
  \caption{Decoding data with unknown style using a Mix-SHAPED model. The data is run through all encoders and decoders. The output of private encoders is fed into a classifier that estimates style distribution. The output symbol distribution is a mixture over all decoder outputs. }
  \label{Fig:mixture-network}
\end{figure}

We treat both of these cases similarly, as a case of modeling an unknown style.
We first describe our treatment of such a case at run-time.
We use a latent random variable $z \in \{1, \ldots, |\mathbb{D}|\}$ to denote the underlying style of a given input.
When generating a token at step $t$, the output token distribution takes the form of a mixture of SHAPED (Mix-SHAPED) model outputs:
\begin{equation}
\begin{split}
p(o_t|\mathbf{x},\mathbf{y}_{1:t-1}) & = \\
 & \hspace{-15mm} \sum_{d=1}^{|\mathbb{D}|} p(o_t|\mathbf{x},\mathbf{y}_{1:t-1}, z=d)p(z=d|\mathbf{x}),
\label{Eq:mix_word}
\end{split}
\end{equation}
where $p(o_t|\mathbf{x},\mathbf{y}_{1:t-1}, z=d)$ is the output symbol distribution of SHAPED decoder $d$, evaluated as in Eq.~\ref{Eq:ProjSoftmax}.
Fig.~\ref{Fig:mixture-network} contains an illustration of such a model.
In this formulation, $p(z|\mathbf{x})$ denotes the style conditional probability distribution from a trainable style classifier.

\begin{figure}[!h]
\centering
\begin{subfigure}[t]{0.45\textwidth}
 \includegraphics[width=\linewidth]{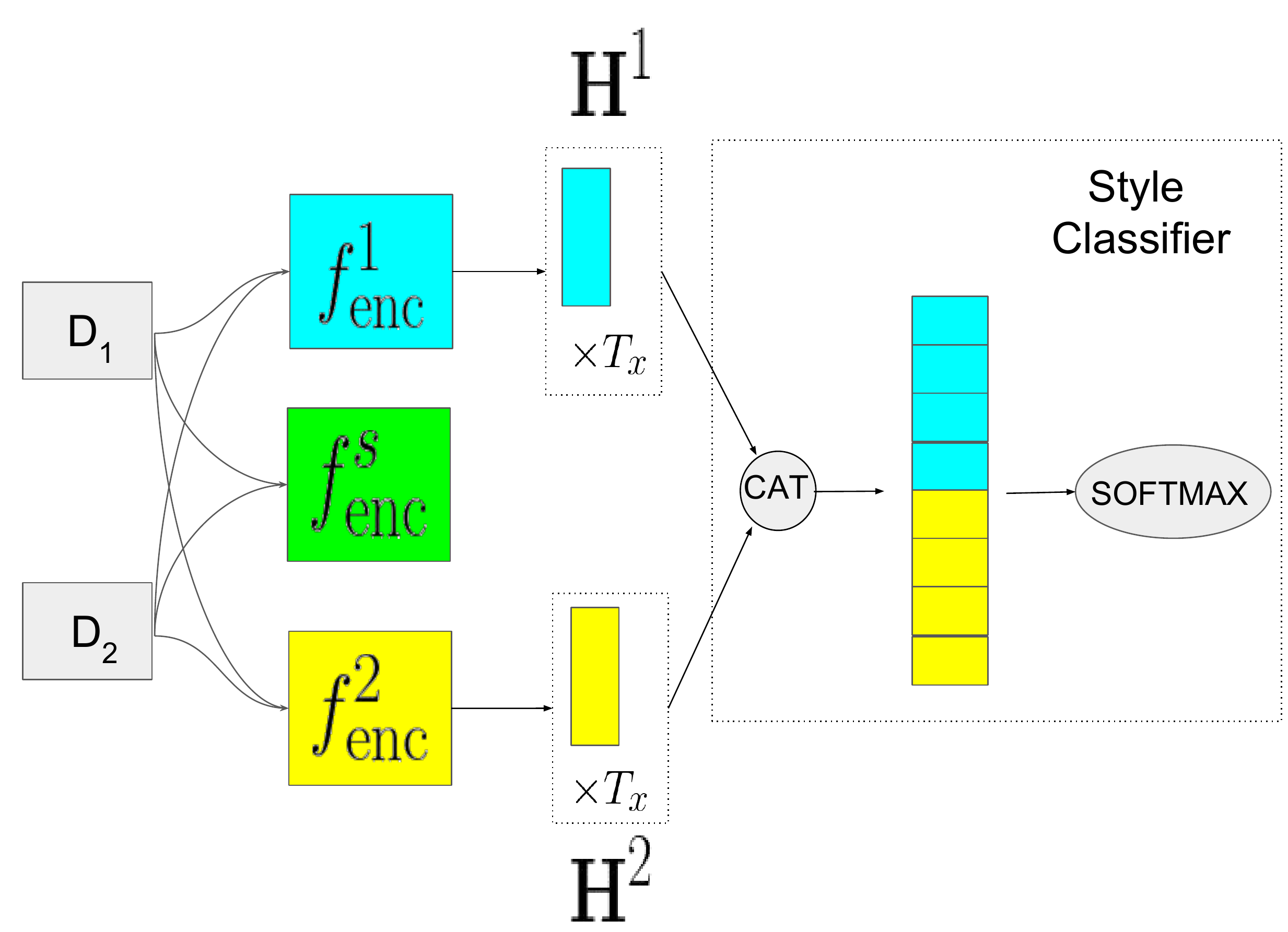}
\end{subfigure}
 \hfill
\begin{subfigure}[t]{0.45\textwidth}
 \includegraphics[width=\linewidth]{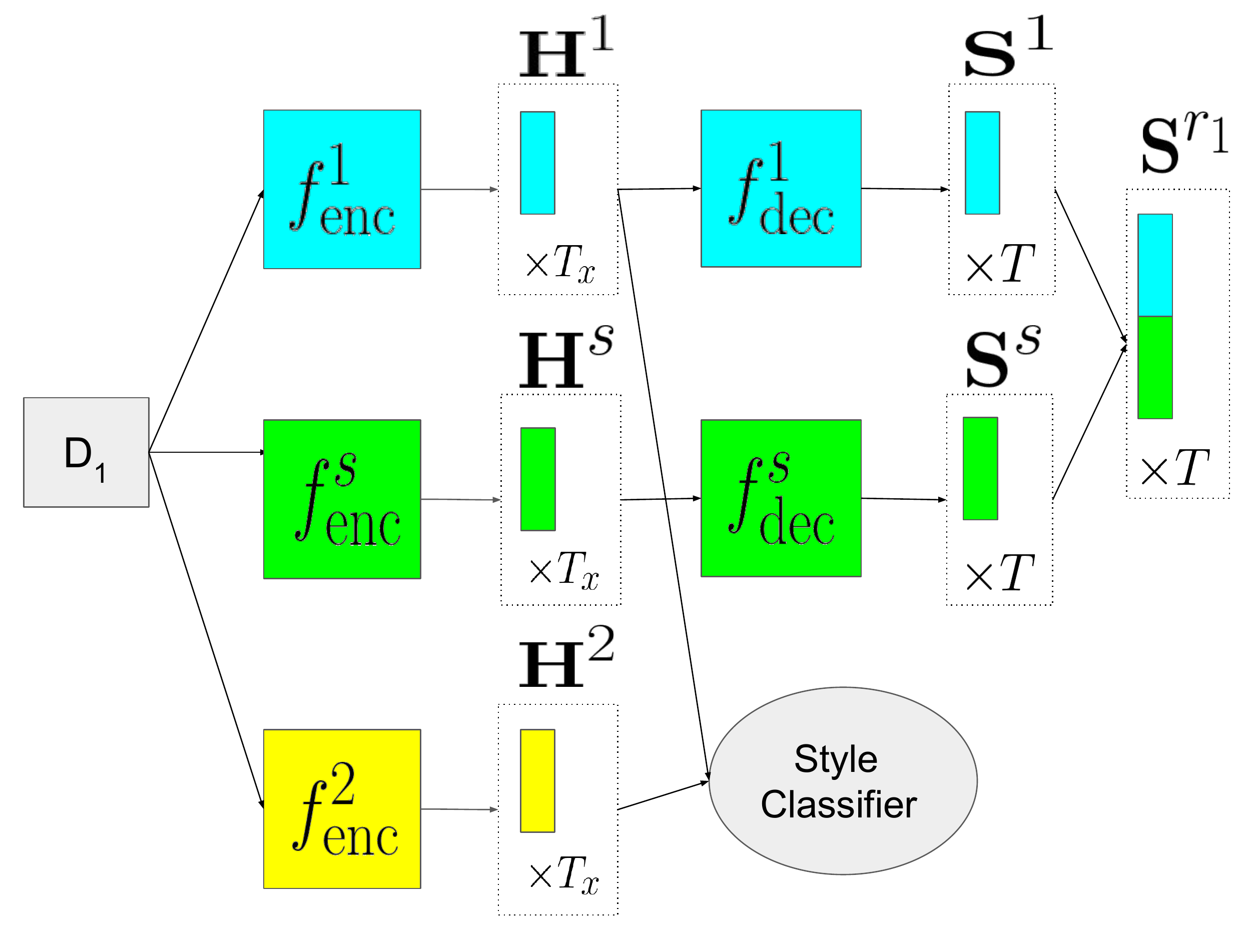}
\end{subfigure}
\caption{Training a Mix-SHAPED model. (a) Each example is fed to all private encoders $f_{enc}^1, f_{enc}^2$, whose outputs are concatenated and fed to a style classifier.  (b) The $D_1$ examples only use $f_{enc}^1, f_{dec}^1, f_{enc}^s, f_{dec}^s$ to decode texts. Private encoder-decoders of other styles are not used. }
\label{fig:MIX_SHAPED_Train}
\end{figure}

The joint data likelihood of target sequence $\mathbf{y}$ and target domain label $z$ for input sequence $\mathbf{x}$ is:
\begin{equation}
p(\mathbf{y},z|\mathbf{x}) = p(\mathbf{y}|z,\mathbf{x})\cdot p(z|\mathbf{x})
\end{equation}
Training the Mix-SHAPED model involves minimizing a loss function that combines the negative log-likelihood of the style labels and the negative log-likelihood of the symbol sequences (see the model in Fig~\ref{fig:MIX_SHAPED_Train}):
\begin{equation}
\begin{split}
\text{Loss}_{\text{Mix-SHAPED}} = -\sum_{i=1}^N\log p(z^{(i)}|\mathbf{x}^{(i)}) \\
-\sum_{i=1}^N
\log p(\mathbf{y}^{(i)}|\mathbf{x}^{(i)},z^{(i)}).
\label{eq:total_loss}
\end{split}
\end{equation}
At run-time, if the style $d$ of the input is available and $d\in\mathbb{D}$, we decode the sequence using Eq.~\ref{Eq:ProjSoftmax}.
This also corresponds to the case $p(z=d|\mathbf{x})=1$ and 0 for all other styles, and reduces Eq.~\ref{Eq:mix_word} to Eq.~\ref{Eq:ProjSoftmax}.
If the style of the input is unknown (or known, but with $d'\not\in\mathbb{D}$), we decode the sequence using Eq.~\ref{Eq:mix_word}, in which case the mixture over SHAPED models given by $p(z|\mathbf{x})$ is approximating the desired output style.

\section{Model Instantiation}
\label{sec:instantiation}
As an implementation of the encoder-decoder model, we use the attention-based sequence-to-sequence model from~\citep{bahdanau2015neural},
with an RNN architecture using GRU units~\citep{chung2014empirical}.
The input token sequences are first projected into an embedding space via an embedding matrix $\mathbf{E}$,
resulting in a sequence of vectors as input representations.

The private and shared RNN cells generate a sequence of hidden state vectors $\mathbf{H}^z=\{h_j^z\}$, $z\in\{1,...,|\mathbb{D}|\}$ and $\mathbf{H}^s=\{h_j^s\}$, for $j\in \{1,...,T_x\}$.
At each step in the encoder, $h_j^z$ and $h_j^s$ are concatenated to form a new output vector $h_j^{r_z}=[h_j^z,h_j^s]$.
The final state of each encoder is used as the initial state of the corresponding decoder.
At time step $t$ in the decoder, the private and shared RNN cell first generate hidden state vectors $\{s^z_t\}, z\in\{1,...,|\mathbb{D}|\}$ and $s^s_t$, then $s^s_t$ is concatenated with each $s^z_t$ to form new vectors $\{s_t^{r_z}\}$ ($z\in\{1,...,|\mathbb{D}|\}$).

We apply the attention mechanism on $s_t^{r_z}$, using attention weights calculated as:
\begin{equation}
q^{r_z}_{tj}=v_a\text{tanh}(W_ah_j^{r_z}+U_as_t^{r_z}),
\end{equation}
which are normalized to a probability distribution:
\begin{equation}
\alpha_{tj}^{r_z} = \frac{\text{exp}(q_{tj}^{r_z})}{\sum_{i=1}^{T_x}\text{exp}(q_{ti}^{r_z})}
\end{equation}
Context vectors are computed using normalized attention weights:
\begin{equation}
  \quad c_t^{r_z} = \sum_{j=1}^{T_x}\alpha^{r_z}_{tj}h_j^{r_z}
\end{equation}
Given the context vector and the hidden state vectors, the symbol distribution at step $t$ is:
\begin{equation}
  p(o_t|\mathbf{x},\mathbf{y}_{1:t}, z) = \text{softmax} (g([c_t^{r_z},s_t^{r_z}]))
  \label{eq:symbol_dist}
\end{equation}
The attention weights in $W_a$, $U_a$, and $v_a$, as well as the embedding matrix $\mathbf{E}$ and vocabulary $V$ are shared by all encoders and decoders.
We use Eq.~\ref{eq:symbol_dist} to calculate the symbol loss in Eq.~\ref{eq:total_loss}.

\section{Quantitative Experiments}
\label{sec:experiment}
We perform a battery of quantitative experiments, designed to answer several main questions:
1) Do the proposed model improve generation performance over alternative approaches?
2) Can a style classifier built using an auxiliary loss provide a reliable estimate on text style?
3) In the case of unknown style, does the Mix-SHAPED model improve generation performance over alternative approaches?
4) To what extent do our models capture style characteristics as opposed to, say, content characteristics?

We perform our experiments using text summarization as the main task.
More precisely, we train and evaluate headline generation models using the publicly-available Gigaword dataset~\citep{graff2003english,napoles2012annotated}.

\subsection{Headline-generation Setup}
The Gigaword dataset contains news articles from seven publishers:
Agence France-Presse (AFP), Associated Press Worldstream (APW), Central News Agency of Taiwan (CNA), Los Angeles Times/Washington Post Newswire Service (LTW), New York Times (NYT), Xinhua News Agency (XIN), and Washington Post/Bloomberg Newswire Service (WPB).
We pre-process this dataset in the same way as in~\cite{rush2015neural}, which results in articles with average length 31.4 words, and headlines with average length 8.5 words.

We consider the publisher identity as a proxy for style, and choose to model as in-domain styles the set $\mathbb{D} = \{$AFP, APW, NYT, XIN$\}$, while holding out CNA and LTW for out-of-domain style testing.
This results in a training set containing the following number of (article, headline) instances: 993,584 AFP, 1,493,758 APW, 578,259 NYT, and 946,322 XIN.
For the test set, we sample a total number of 10,000 in-domain examples from the original Gigawords test dataset, which include 2,886 AFP, 2,832 APW, 1,610 NYT, and 2,012 XIN.
For out-of-domain testing, we randomly sample 10,000 LTW and 10,000 CNA test data examples.
We remove the WPB articles due to their small number of instances.

\subsubsection{Experimental Setup}
We compare the following models:
\begin{itemize}
\item A Shared encoder-decoder model (S) trained on all styles in $\mathbb{D}$;
\item A suite of Private encoder-decoder models (P), each one trained on a particular style from $\mathbb{D}=\{$AFP, APW, NYT, XIN$\}$;\footnote{We also tried to warm-start a private model using the best checkpoint of the shared model, but found that it cannot improve over the shared model.}
\item A SHAPED model (SP) trained on all styles in $\mathbb{D}$; at test time, the style of test data is provided to the model; the article is only run through its style-specific private network and shared network (style classifier is not needed);
\item A Mix-SHAPED model (M-SP) trained on all styles in $\mathbb{D}$; at test time, the style of article is not provided to the model; the output is computed using the mixture model, with the estimated style probabilities from the style classifier used as weights.
\end{itemize}
When testing on the out-of-domain styles CNA/LTW, we only compare the Shared (S) model with the Mix-SHAPED (M-SP) model, as the others cannot properly handle this scenario.

As hyper-parameters for the model instantiation, we used 500-dimension word embeddings, and a three-layer, 500-dimension GRU-cell RNN architecture; the encoder was instantiated as a bi-directional RNN.
The lengths of the input and output sequences were truncated to 40 and 20 tokens, respectively.
All the models were optimized using Adagrad~\citep{duchi2011adaptive}, with an initial learning rate of 0.01.
The training procedure was done over mini-batches of size 128, and the updates were done asynchronously across 40 workers for 5M steps.
The encoder/decoder word embedding and the output projection matrices were tied to minimize the number of parameters.
To avoid the slowness from the softmax operator over large vocabulary sizes, and also mitigate the impact of out-of-vocabulary tokens, we applied a subtokenization method~\citep{wu2016gnmt}, which invertibly transforms a native token into a sequence of subtokens from a limited vocabulary (here set to 32K).

\paragraph{Comparison with Previous Work}
In the next section, we report our main results using the in-domain and out-of-domain (w.r.t. the selected publisher styles) test sets described above, since these test sets have a balanced publisher style frequency that allows us to measure the impact of our style-adaptation models.
However, we also report here the performance of our Shared (S) baseline model (with the above hyper-parameters) on the original 2K test set used in~\cite{rush2015neural}.
On that test set, our S model obtains 30.13 F1 ROUGE-L score, compared to 28.34 ROUGE-L obtained by the ABS+ model~\cite{rush2015neural}, and 30.64 ROUGE-L obtained by the words-lvt2k-1sent model~\cite{nallapati2016abstractive}.
This comparison indicates that our S model is a competitive baseline, making the comparisons against the SP and M-SP models meaningful when using our in-domain and out-of-domain test sets.

\subsubsection{Main Results}
\begin{figure}[h]
  \centering
  \begin{subfigure}{0.45\textwidth}
\includegraphics[width=\textwidth]{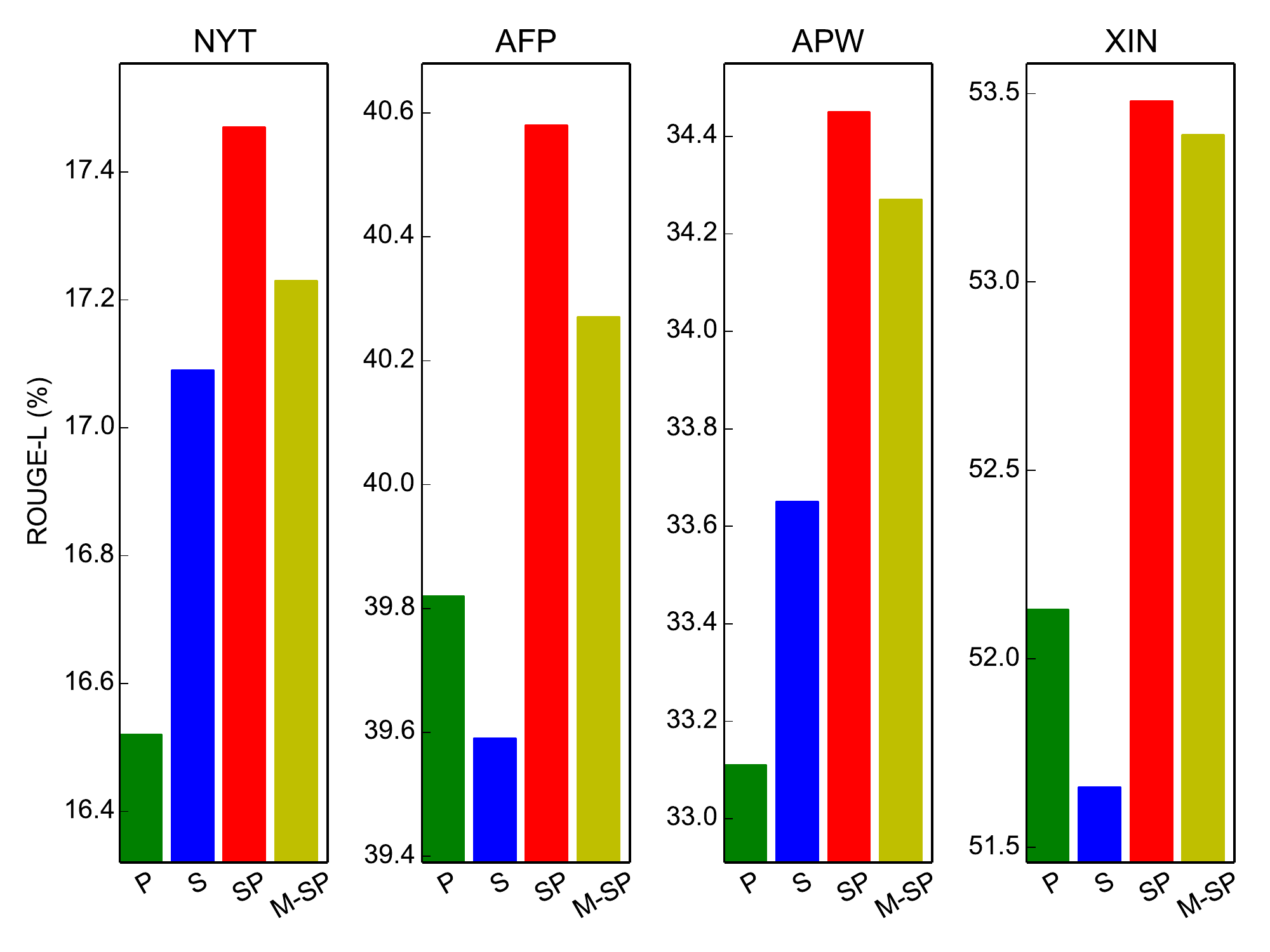}
  \caption{Rouge-L scores on headline generation, shown separately on four in-domain styles.}
  \label{fig:four_domain_scores}
\end{subfigure}
\hfill
\begin{subfigure}{0.45\textwidth}
\includegraphics[width=\textwidth]{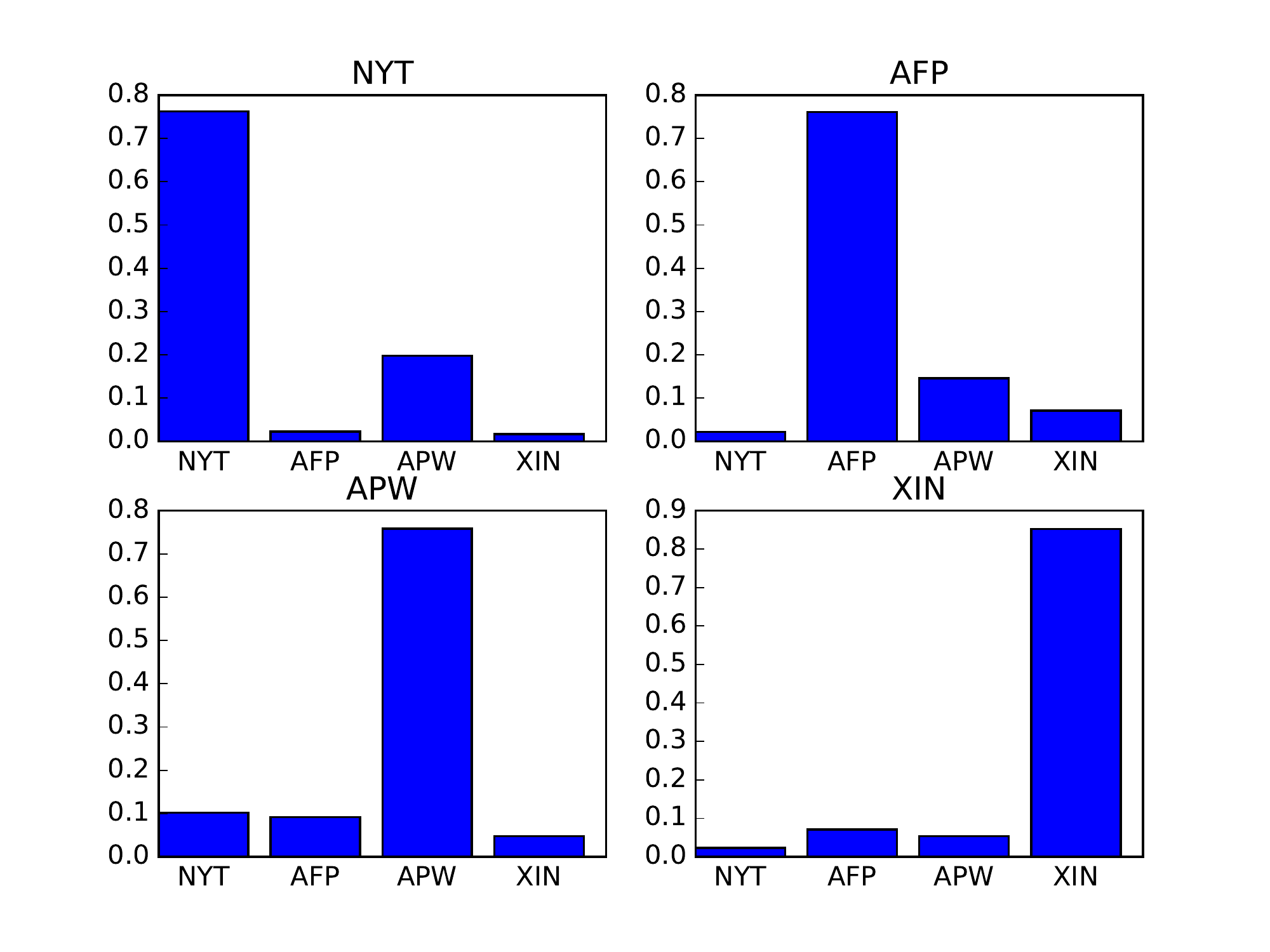}
  \caption{Average estimated probability distribution by the M-SP model over the four styles, for each in-domain target style in the test set.}
  \label{fig:four_domain_probs}
\end{subfigure}
\caption{Experimental results on the headline generation task, for in-domain styles.}
\end{figure}

The Rouge scores for the in-domain testing data are reported in Table \ref{table:scores_all_four_domains} (over the combined AFP/APW/XIN/NYT testset) and Fig.~\ref{fig:four_domain_scores} (over individual-style test sets).
The numbers indicate that the SP and M-SP models consistently outperform the S and P model, supporting the conclusion that the S model loses important characteristics due to averaging effects, while the P models miss the opportunity to efficiently exploit the training data.
Additionally, the performance of SP is consistently better than M-SP in this setting, which indicates that the style label is helpful.
As shown in Fig. \ref{fig:four_domain_probs}, the style classifier achieves around 80\% accuracy overall in predicting the style under the M-SP model, with some styles (e.g., XIN) being easier to predict than others.
The performance of the classifier is directly reflected in the quantitative difference between the SP and M-SP models on individual-style test sets (see Fig.~\ref{fig:four_domain_scores}, where the XIN style has the smallest difference between the two models).

\begin{table}
\small
\centering
\begin{tabular}{ |c|c|c|c|}
\hline
&\multicolumn{3}{|c|}{AFP/APW/XIN/NYT Test}  \\
\hline
&Rouge-1 & Rouge-2 & Rouge-L\\
\hline
P & 39.14\scriptsize{$\pm$0.47} & 19.74\scriptsize{$\pm$0.48} & 36.42\scriptsize{$\pm$0.46} \\
\hline
S & 39.32\scriptsize{$\pm$0.26} & 19.63\scriptsize{$\pm$0.24} & 36.51\scriptsize{$\pm$0.26} \\
\hline
SP & \textbf{40.34}\scriptsize{$\pm$0.26} & \textbf{20.38}\scriptsize{$\pm$0.25} & \textbf{37.52}\scriptsize{$\pm$0.25}\\
\hline
M-SP & 40.10\scriptsize{$\pm$0.25} & 20.21\scriptsize{$\pm$0.26} & 37.30\scriptsize{$\pm$0.26}\\
\hline
\end{tabular}
\caption{ROUGE F1 scores on the combined AFP/APW/XIN/NYT in-domain test set. }
\label{table:scores_all_four_domains}
\end{table}

\begin{figure}
\includegraphics[width=0.45\textwidth]{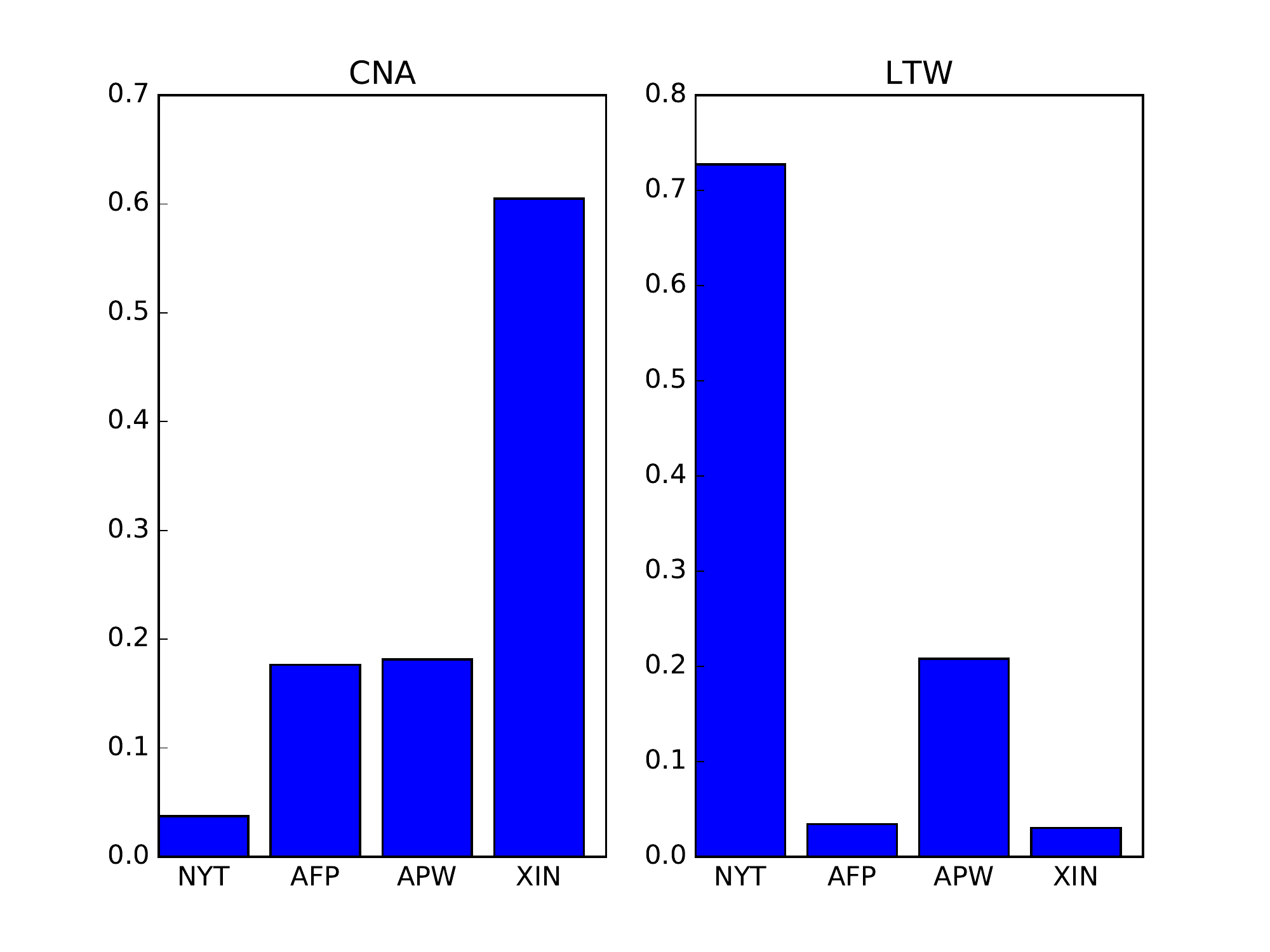}
\caption{Estimated style probabilities over the four in-domain styles AFP/APW/XIN/NYT, for out-of-domain styles CNA and LTW.}
\label{fig:probs_CNA_LTW}
\end{figure}

The evaluation results for the out-of-domain scenario are reported in Table~\ref{table:CNA_LTW}.
The numbers indicate that the M-SP model significantly outperforms the S model, supporting the conclusion that the M-SP model is capable of performing on-the-fly adaptation of output style.
This conclusion is further strengthened by the style probability distributions shown in Fig~\ref{fig:probs_CNA_LTW}:
they indicate that, for the out-of-domain CNA style, the output mixture is heavily weighted towards the XIN style (0.6 of the probability mass), while for the LTW style, the output mixture weights heavily the NYT style (0.72 of the probability mass).
This result is likely to reflect true style characteristics shared by these publishers, since both CNA and XIN are produced by Chinese news agencies (from Taiwan and mainland China, respectively), while both LTW and NYT are U.S. news agencies owned by the same media corporation.

\begin{table*}[!htb]
\centering
\begin{tabular}{ |c|c|c|c|c|c|c|}
 \hline
& \multicolumn{3}{|c|}{CNA Test} & \multicolumn{3}{|c|}{LTW Test} \\
 \hline
  & Rouge-1 & Rouge-2 & Rouge-L & Rouge-1 & Rouge-2 & Rouge-L \\
 \hline
S &  40.73\scriptsize{$\pm$0.21} & 17.75\scriptsize{$\pm$0.18} & 37.70\scriptsize{$\pm$0.20} & 27.08\scriptsize{$\pm$0.19} & 8.97\scriptsize{$\pm$0.15} & 25.01\scriptsize{$\pm$0.17} \\
\hline
M-SP &  \textbf{42.00}\scriptsize{$\pm$0.20} & \textbf{19.48}\scriptsize{$\pm$0.21} & \textbf{39.24}\scriptsize{$\pm$0.22}& \textbf{27.79}\scriptsize{$\pm$0.19}&  \textbf{9.31}\scriptsize{$\pm$0.18}& \textbf{25.60}\scriptsize{$\pm$0.17} \\
\hline
\end{tabular}
\caption{ROUGE F1 scores on out-of-domain style test sets CNA and LTW. }
\label{table:CNA_LTW}
\end{table*}

\subsubsection{Experiment Variants}

\paragraph{Model capacity}
In order to remove the possibility that the improved performance of the SP model is due simply to an increased model size compared to the S model, we perform an experiment in which we triple the size of the GRU cell dimensions for the S model.
However, we find no significant performance difference compared to the original dimensions (the ROUGE-L score of the triple-size S model is 36.61, compared to 36.51 obtained of the original S model).

\paragraph{Style embedding}
A competitive approach to modeling different styles is to directly encode the style information into the embedding space. In~\citep{johnson2016enabling}, the style label is converted into a one-hot vector and is concatenated with the word embedding at each time step in the S model. The outputs of this model are at 36.68 ROUGE-L, slightly higher than the baseline S model, but significantly lower than the SP model performance (37.52 ROUGE-L).

Another style embedding approach is to augment the S model with continuous trainable style embeddings for each predefined style label, similar to~\citep{ammarMBDS16}. The resulting outputs achieve 37.2 ROUGE-L, which is better than the S model with one-hot style embedding, but still worse than the SP method (statistically significant at p-value=0.025 using paired t-test). However, neither of these approaches apply to the cases when the style is  out-of-domain or unknown during testing. In contrast, such cases are handled naturally by the proposed M-SP model.

\paragraph{Ensemble model}
Another question is whether the SP model simply benefits from ensembling multiple models rather than style adaptation. To answer this question, we apply a uniform mixture over the private model output along with the shared model output, rather than using the learnt probability distribution from the style classifier. The ROUGE-1/2/L scores are 39.9/19.7/37.0. They are higher than the S model but significantly lower than the SP model and the M-SP model (p-value 0.016). This result confirms that the information that the style classifier encodes is beneficiary, and leads to improved performance.

\paragraph{Style vs. Content}
Previous experiments indicate that the SP and M-SP models have superior generation accuracy, but it is unclear to what extent the difference comes from improved modeling of style versus modeling of content.
To clarify this issue, we performed an experiment in which we replace the named entities appearing in both article and headline with corresponding entity tags, in effect suppressing almost completely any content signal.
For instance, given an input such as ``China called Thursday on the parties involved in talks on North Korea's nuclear program to show flexibility as a deadline for implementing the first steps of a breakthrough deal approached.", paired with goldtruth output ``China urges flexibility as NKorea deadline approaches", we replaced the named entities with their types, and obtained: ``LOC\_0 called Thursday on the ORG\_0 involved in NON\_2 on LOC\_1 's NON\_3 to show NON\_0 as a NON\_1 for implementing the first NON\_4 of a NON\_5 approached .", paired with ``LOC\_0 urges NON\_0 as LOC\_1 NON\_1 approaches."

Under this experimental conditions, both the SP and M-SP models still achieve significantly better performance compared to the S baseline.
On the combined AFP/APW/XIN/NYT in-domain test set, the SP model achieves 61.70 ROUGE-L and M-SP achieves 61.52 ROUGE-L, compared to 60.20 ROUGE-L obtained by the S model. On the CNA/LTW out-of-domain test set, M-SP achieves 60.75 ROUGE-L, compared to 59.47 ROUGE-L by the S model.

\begin{table*}[!htbp]
\centering
\begin{tabular}{|c|p{12cm}|}
 \hline
 article & the org\_2 is to forge non\_1 with the org\_3 located in loc\_2 , loc\_1 , the per\_0 of the loc\_0 org\_4 said tuesday . \\
 \hline
 title & loc\_0 org\_0 to forge non\_0 with loc\_1 org\_1\\
 \hline
 output by S & org\_0 to org\_1 in non\_0 \\
 \hline
 output by M-SP & loc\_0 org\_0 to forge non\_0 with loc\_1 org\_1\\
 \hline\hline
 article & loc\_0 - born per\_0 per\_0 will pay non\_1 here next month to per\_1 , the org\_2 ( org\_1 ) per\_1 who per\_1 perished in an non\_2 in february , the org\_3 said thursday . \\
 \hline
 title & per\_0 to pay non\_0 to late org\_1 org\_0\\
 \hline
 output by S & per\_0 to visit org\_0 in non\_0  \\
 \hline
 output by M-SP & per\_0 to pay non\_0 to org\_1 org\_0\\
 \hline
\end{tabular}
\caption{Examples of input article (and groundtruth title) and output generated by S and M-SP. Named entities in the training instances (both article and title) are replaced the entity type. }
\label{table:style_example}
\end{table*}

In Table~\ref{table:style_example}, we show an example which indicates the ability of style adaptation benifiting summarization.
For instance, we find that both CNA and XIN make more frequent use of the style pattern ``xxx will/to [verb] yyy $\ldots$, zzz said ???day" (about 15\% of CNA articles contain this pattern, while only 2\% of LTW articles have it).
From Table~\ref{table:style_example}, we can see that the S model sometimes misses or misuses the verb in its output, while the M-SP model does a much better job at capturing both the verb/action as well as other relations (via prepositions, etc.)

\begin{figure}[!htbp]
\includegraphics[width=0.45\textwidth]{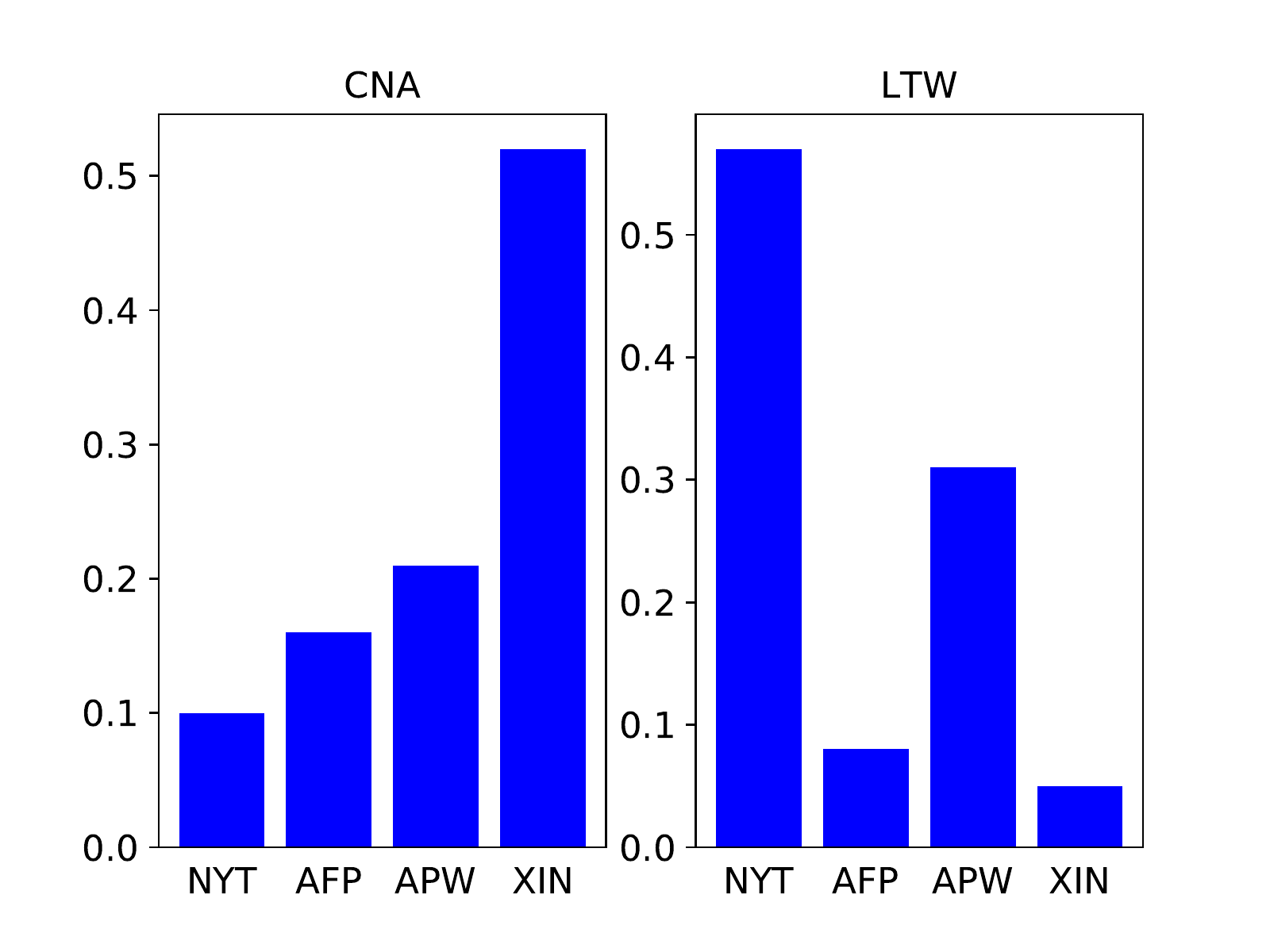}
\caption{Estimated style probabilities over the four in-domain styles AFP/APW/XIN/NYT, for out-of-domain styles CNA and LTW, after  named entities in the article and summary are replaced with entity tags. }
\label{fig:probs_CNA_LTW_entities}
\end{figure}

Fig.~\ref{fig:probs_CNA_LTW_entities} shows the estimated style probabilities over the four styles AFP/APW/XIN/NYT for CNA and LTW, under this experiment condition.
We observe that, in this version as well, CNA is closely matching the style of XIN, while LTW is matching that of NYT.
The distribution is similar to the one in Fig.~\ref{fig:probs_CNA_LTW}, albeit a bit flatter as a result of content removal.
As such, it supports the conclusion that the classifier indeed learns style (in addition to content) characteristics.

\section{Conclusion}
\label{sec:conclusion}
In this paper, we describe two new style-adaptation model architectures for text sequence generation tasks, SHAPED and Mix-SHAPED.
Both versions are shown to significantly outperform models that are either trained in a manner that ignores style characteristics (and hence exhibit a style-averaging effect in their outputs), or models that are trained single-style.

The latter is a particularly interesting result, as a model that is trained (with enough data) on a single-style and evaluated on the same style would be expected to exhibit the highest performance.
Our results show that, even for single-style models trained on over 1M examples, their performance is inferior to the performance of SHAPED models on that particular style.

Our conclusion is that the proposed architectures are both efficient and effective in modeling both generic language phenomena, as well as particular style characteristics, and are capable of producing higher-quality abstractive outputs that take into account style characteristics.

\clearpage
\bibliography{garcon}

\begin{thebibliography}{}
\expandafter\ifx\csname natexlab\endcsname\relax\def\natexlab#1{#1}\fi

\bibitem[{Ammar et~al.(2016)Ammar, Mulcaire, Ballesteros, Dyer, and
  Smith}]{ammarMBDS16}
Waleed Ammar, George Mulcaire, Miguel Ballesteros, Chris Dyer, and Noah~A.
  Smith. 2016.
\newblock Many languages, one parser.
\newblock {\em {TACL}\/} 4:431--444.

\bibitem[{Asghar et~al.(2017)Asghar, Poupart, Jiang, and Li}]{asghar2017deep}
Nabiha Asghar, Pascal Poupart, Xin Jiang, and Hang Li. 2017.
\newblock Deep active learning for dialogue generation.
\newblock In {\em Proceedings of the 6th Joint Conference on Lexical and
  Computational Semantics (*SEM 2017)\/}. pages 78--83.

\bibitem[{Bahdanau et~al.(2015)Bahdanau, Cho, and Bengio}]{bahdanau2015neural}
D.~Bahdanau, K.~Cho, and Y.~Bengio. 2015.
\newblock Neural machine translation by jointly learning to align and
  translate.
\newblock In {\em Proceedings of ICLR\/}.

\bibitem[{Bousmalis et~al.(2016)Bousmalis, Trigeorgis, Silberman, Krishnan, and
  Erhan}]{bousmalis2016domain}
Konstantinos Bousmalis, George Trigeorgis, Nathan Silberman, Dilip Krishnan,
  and Dumitru Erhan. 2016.
\newblock Domain separation networks.
\newblock In {\em Advances in Neural Information Processing Systems\/}. pages
  343--351.

\bibitem[{Chopra et~al.(2016)Chopra, Auli, Rush, and
  Harvard}]{chopra2016abstractive}
Sumit Chopra, Michael Auli, Alexander~M Rush, and SEAS Harvard. 2016.
\newblock Abstractive sentence summarization with attentive recurrent neural
  networks.
\newblock In {\em HLT-NAACL\/}. pages 93--98.

\bibitem[{Chopra et~al.(2013)Chopra, Balakrishnan, and
  Gopalan}]{chopra2013dlid}
Sumit Chopra, Suhrid Balakrishnan, and Raghuraman Gopalan. 2013.
\newblock Dlid: Deep learning for domain adaptation by interpolating between
  domains.
\newblock In {\em ICML workshop on challenges in representation learning\/}.
  volume~2.

\bibitem[{Chung et~al.(2014)Chung, Gulcehre, Cho, and
  Bengio}]{chung2014empirical}
Junyoung Chung, Caglar Gulcehre, KyungHyun Cho, and Yoshua Bengio. 2014.
\newblock Empirical evaluation of gated recurrent neural networks on sequence
  modeling.
\newblock {\em arXiv preprint arXiv:1412.3555\/} .

\bibitem[{Daum{\'e}~III(2009)}]{daume2009frustratingly}
Hal Daum{\'e}~III. 2009.
\newblock Frustratingly easy domain adaptation.
\newblock {\em arXiv preprint arXiv:0907.1815\/} .

\bibitem[{Duchi et~al.(2011)Duchi, Hazan, and Singer}]{duchi2011adaptive}
John Duchi, Elad Hazan, and Yoram Singer. 2011.
\newblock Adaptive subgradient methods for online learning and stochastic
  optimization.
\newblock {\em Journal of Machine Learning Research\/} 12(Jul):2121--2159.

\bibitem[{Filippova et~al.(2015)Filippova, Alfonseca, Colmenares, Kaiser, and
  Vinyals}]{filippova2015sentence}
Katja Filippova, Enrique Alfonseca, Carlos Colmenares, Lukasz Kaiser, and Oriol
  Vinyals. 2015.
\newblock Sentence compression by deletion with lstms.
\newblock In {\em Proceedings of the 2015 Conference on Empirical Methods in
  Natural Language Processing (EMNLP'15)\/}.

\bibitem[{Glorot et~al.(2011)Glorot, Bordes, and Bengio}]{glorot2011domain}
Xavier Glorot, Antoine Bordes, and Yoshua Bengio. 2011.
\newblock Domain adaptation for large-scale sentiment classification: A deep
  learning approach.
\newblock In {\em Proceedings of the 28th international conference on machine
  learning (ICML-11)\/}. pages 513--520.

\bibitem[{Graff and Cieri(2003)}]{graff2003english}
David Graff and C~Cieri. 2003.
\newblock English gigaword corpus.
\newblock {\em Linguistic Data Consortium\/} .

\bibitem[{Hochreiter and Schmidhuber(1997)}]{hochreiter1997long}
Sepp Hochreiter and J{\"u}rgen Schmidhuber. 1997.
\newblock Long short-term memory.
\newblock {\em Neural computation\/} 9(8):1735--1780.

\bibitem[{Hua and Wang(2017)}]{hua2017pilot}
Xinyu Hua and Lu~Wang. 2017.
\newblock A pilot study of domain adaptation effect for neural abstractive
  summarization.
\newblock {\em arXiv preprint arXiv:1707.07062\/} .

\bibitem[{Johnson et~al.(2016)Johnson, Schuster, Le, Krikun, Wu, Chen, Thorat,
  Vi{\'{e}}gas, Wattenberg, Corrado, Hughes, and Dean}]{johnson2016enabling}
Melvin Johnson, Mike Schuster, Quoc~V. Le, Maxim Krikun, Yonghui Wu, Zhifeng
  Chen, Nikhil Thorat, Fernanda~B. Vi{\'{e}}gas, Martin Wattenberg, Greg
  Corrado, Macduff Hughes, and Jeffrey Dean. 2016.
\newblock \href{http://arxiv.org/abs/1611.04558}{Google's multilingual neural
  machine translation system: Enabling zero-shot translation}.
\newblock {\em CoRR\/} abs/1611.04558.
\newblock \url{http://arxiv.org/abs/1611.04558}.

\bibitem[{Kim et~al.(2016)Kim, Stratos, and Sarikaya}]{kim2016frustratingly}
Young-Bum Kim, Karl Stratos, and Ruhi Sarikaya. 2016.
\newblock Frustratingly easy neural domain adaptation.
\newblock In {\em Proceedings of the 26th International Conference on
  Computational Linguistics (COLING)\/}.

\bibitem[{Li et~al.(2016)Li, Monroe, Ritter, Galley, Gao, and
  Jurafsky}]{li2016deep}
Jiwei Li, Will Monroe, Alan Ritter, Michel Galley, Jianfeng Gao, and Dan
  Jurafsky. 2016.
\newblock Deep reinforcement learning for dialogue generation.
\newblock {\em arXiv preprint arXiv:1606.01541\/} .

\bibitem[{Li et~al.(2017)Li, Monroe, Shi, Ritter, and
  Jurafsky}]{li2017adversarial}
Jiwei Li, Will Monroe, Tianlin Shi, Alan Ritter, and Dan Jurafsky. 2017.
\newblock Adversarial learning for neural dialogue generation.
\newblock {\em arXiv preprint arXiv:1701.06547\/} .

\bibitem[{Liu et~al.(2017)Liu, Zhu, Ye, Guadarrama, and
  Murphy}]{liu2017optimization}
Siqi Liu, Zhenhai Zhu, Ning Ye, Sergio Guadarrama, and Kevin Murphy. 2017.
\newblock Optimization of image description metrics using policy gradient
  methods.
\newblock In {\em International Conference on Computer Vision (ICCV)\/}.

\bibitem[{Long et~al.(2016)Long, Zhu, Wang, and Jordan}]{long2016unsupervised}
Mingsheng Long, Han Zhu, Jianmin Wang, and Michael~I Jordan. 2016.
\newblock Unsupervised domain adaptation with residual transfer networks.
\newblock In {\em Advances in Neural Information Processing Systems\/}. pages
  136--144.

\bibitem[{Nallapati et~al.(2016)Nallapati, Zhou, Gulcehre, Xiang
  et~al.}]{nallapati2016abstractive}
Ramesh Nallapati, Bowen Zhou, Caglar Gulcehre, Bing Xiang, et~al. 2016.
\newblock Abstractive text summarization using sequence-to-sequence {RNN}s and
  beyond.
\newblock In {\em Proceedings of CoNLL\/}.

\bibitem[{Napoles et~al.(2012)Napoles, Gormley, and
  Van~Durme}]{napoles2012annotated}
Courtney Napoles, Matthew Gormley, and Benjamin Van~Durme. 2012.
\newblock Annotated gigaword.
\newblock In {\em Proceedings of the Joint Workshop on Automatic Knowledge Base
  Construction and Web-scale Knowledge Extraction\/}. Association for
  Computational Linguistics, pages 95--100.

\bibitem[{Nema et~al.(2017)Nema, Khapra, Laha, and
  Ravindran}]{nema2017diversity}
Preksha Nema, Mitesh Khapra, Anirban Laha, and Balaraman Ravindran. 2017.
\newblock Diversity driven attention model for query-based abstractive
  summarization.
\newblock {\em arXiv preprint arXiv:1704.08300\/} .

\bibitem[{Paulus et~al.(2017)Paulus, Xiong, and Socher}]{paulus2017deep}
Romain Paulus, Caiming Xiong, and Richard Socher. 2017.
\newblock A deep reinforced model for abstractive summarization.
\newblock {\em arXiv preprint arXiv:1705.04304\/} .

\bibitem[{Peng and Dredze(2017)}]{peng2017multi}
Nanyun Peng and Mark Dredze. 2017.
\newblock Multi-task domain adaptation for sequence tagging.
\newblock {\em ACL 2017\/} page~91.

\bibitem[{Ranzato et~al.(2015)Ranzato, Chopra, Auli, and Zaremba}]{mixer15}
Marc'Aurelio Ranzato, Sumit Chopra, Michael Auli, and Wojciech Zaremba. 2015.
\newblock Sequence level training with recurrent neural networks.
\newblock {\em CoRR\/} abs/1511.06732.

\bibitem[{Rush et~al.(2015)Rush, Chopra, and Weston}]{rush2015neural}
Alexander~M. Rush, Sumit Chopra, and Jason Weston. 2015.
\newblock A neural attention model for abstractive sentence summarization.
\newblock In {\em Proceedings of EMNLP\/}. pages 379--389.

\bibitem[{See et~al.(2017)See, Liu, and Manning}]{see2017get}
Abigail See, Peter~J Liu, and Christopher~D Manning. 2017.
\newblock Get to the point: Summarization with pointer-generator networks.
\newblock In {\em Proceedings of ACL\/}.

\bibitem[{Sener et~al.(2016)Sener, Song, Saxena, and
  Savarese}]{sener2016learning}
Ozan Sener, Hyun~Oh Song, Ashutosh Saxena, and Silvio Savarese. 2016.
\newblock Learning transferrable representations for unsupervised domain
  adaptation.
\newblock In {\em Advances in Neural Information Processing Systems\/}. pages
  2110--2118.

\bibitem[{Sutskever et~al.(2014)Sutskever, Vinyals, and
  Le}]{sutskever2014sequence}
Ilya Sutskever, Oriol Vinyals, and Quoc~V Le. 2014.
\newblock Sequence to sequence learning with neural networks.
\newblock In {\em Advances in neural information processing systems\/}. pages
  3104--3112.

\bibitem[{Tzeng et~al.(2015)Tzeng, Hoffman, Darrell, and
  Saenko}]{tzeng2015simultaneous}
Eric Tzeng, Judy Hoffman, Trevor Darrell, and Kate Saenko. 2015.
\newblock Simultaneous deep transfer across domains and tasks.
\newblock In {\em Proceedings of the IEEE International Conference on Computer
  Vision\/}. pages 4068--4076.

\bibitem[{Vaswani et~al.(2017)Vaswani, Shazeer, Parmar, Uszkoreit, Jones,
  Gomez, Kaiser, and Polosukhin}]{vaswani2017attention}
Ashish Vaswani, Noam Shazeer, Niki Parmar, Jakob Uszkoreit, Llion Jones,
  Aidan~N Gomez, Lukasz Kaiser, and Illia Polosukhin. 2017.
\newblock Attention is all you need.
\newblock In {\em Advances in Neural Information Processing Systems\/}.

\bibitem[{Vinyals et~al.(2015)Vinyals, Toshev, Bengio, and
  Erhan}]{vinyals2015show}
Oriol Vinyals, Alexander Toshev, Samy Bengio, and Dumitru Erhan. 2015.
\newblock Show and tell: A neural image caption generator.
\newblock In {\em Proceedings of the IEEE conference on computer vision and
  pattern recognition\/}. pages 3156--3164.

\bibitem[{Wu et~al.(2016)Wu, Schuster, Chen, Le, Norouzi, Macherey, Krikun,
  Cao, Gao, Macherey, Klingner, Shah, Johnson, Liu, Łukasz Kaiser, Gouws,
  Kato, Kudo, Kazawa, Stevens, Kurian, Patil, Wang, Young, Smith, Riesa,
  Rudnick, Vinyals, Corrado, Hughes, and Dean}]{wu2016gnmt}
Yonghui Wu, Mike Schuster, Zhifeng Chen, Quoc~V. Le, Mohammad Norouzi, Wolfgang
  Macherey, Maxim Krikun, Yuan Cao, Qin Gao, Klaus Macherey, Jeff Klingner,
  Apurva Shah, Melvin Johnson, Xiaobing Liu, Łukasz Kaiser, Stephan Gouws,
  Yoshikiyo Kato, Taku Kudo, Hideto Kazawa, Keith Stevens, George Kurian,
  Nishant Patil, Wei Wang, Cliff Young, Jason Smith, Jason Riesa, Alex Rudnick,
  Oriol Vinyals, Greg Corrado, Macduff Hughes, and Jeffrey Dean. 2016.
\newblock \href{http://arxiv.org/abs/1609.08144}{Google's neural machine
  translation system: Bridging the gap between human and machine translation}.
\newblock {\em CoRR\/} abs/1609.08144.
\newblock \url{http://arxiv.org/abs/1609.08144}.

\bibitem[{Xu et~al.(2015)Xu, Ba, Kiros, Courville, Salakhutdinov, Zemel, and
  Bengio}]{xu2015show}
Kelvin Xu, Jimmy Ba, Ryan Kiros, Aaron Courville, Ruslan Salakhutdinov, Richard
  Zemel, and Yoshua Bengio. 2015.
\newblock Show, attend and tell: Neural image caption generation with visual
  attention.
\newblock In {\em Proc. of the 32nd International Conference on Machine
  Learning (ICML)\/}.

\bibitem[{Zhang and Lapata(2017)}]{zhang2017sentence}
Xingxing Zhang and Mirella Lapata. 2017.
\newblock Sentence simplification with deep reinforcement learning.
\newblock {\em arXiv preprint arXiv:1703.10931\/} .

\bibitem[{Zhou et~al.(2016)Zhou, Xie, Huang, and He}]{zhou2016bi}
Guangyou Zhou, Zhiwen Xie, Jimmy~Xiangji Huang, and Tingting He. 2016.
\newblock Bi-transferring deep neural networks for domain adaptation.
\newblock In {\em ACL (1)\/}.

\end{thebibliography}
\bibliographystyle{acl_natbib}

\end{document}